\documentclass[sigconf,nonacm,screen]{acmart}

\copyrightyear{2025}
\acmYear{2025}
\setcopyright{acmlicensed}
\acmConference[MM '25] {Proceedings of the 33rd ACM International Conference on Multimedia}{October 27--31, 2025}{Dublin, Ireland.}
\acmBooktitle{Proceedings of the 33rd ACM International Conference on Multimedia (MM '25), October 27--31, 2025, Dublin, Ireland}
\acmISBN{979-8-4007-2035-2/2025/10}
\acmDOI{10.1145/3746027.3755420}

\acmSubmissionID{3678}

\usepackage{array}
\usepackage{pifont}
\usepackage[capitalize]{cleveref}
\usepackage{tcolorbox}
\usepackage{fancyvrb}
\usepackage{multirow}
\usepackage{subcaption}
\usepackage{adjustbox}

\newcommand{\etal}{\textit{et al}.}
\newcommand{\ie}{\textit{i}.\textit{e}.}
\newcommand{\eg}{\textit{e}.\textit{g}.}
\newcolumntype{L}[1]{>{\arraybackslash}m{#1}}
\newcolumntype{C}[1]{>{\centering\arraybackslash}m{#1}}
\newcolumntype{R}[1]{>{\raggedleft\arraybackslash}p{#1}}
\newcommand{\todo}[1]{}
\definecolor{spec}{HTML}{B1400D}
\definecolor{img}{HTML}{12711C}
\definecolor{fim}{HTML}{9F3DD3}

\begin{document}

\title{Multimodal Markup Document Models for Graphic Design Completion}

\author{Kotaro Kikuchi}
\email{kikuchi_kotaro_xa@cyberagent.co.jp}
\affiliation{%
  \institution{CyberAgent}
  \city{Shibuya-ku}
  \state{Tokyo}
  \country{Japan}
}

\author{Ukyo Honda}
\email{honda_ukyo@cyberagent.co.jp}
\affiliation{%
  \institution{CyberAgent}
  \city{Shibuya-ku}
  \state{Tokyo}
  \country{Japan}
}

\author{Naoto Inoue}
\email{naoto.inoue.0804@gmail.com}
\affiliation{%
  \institution{CyberAgent}
  \city{Shibuya-ku}
  \state{Tokyo}
  \country{Japan}
}

\author{Mayu Otani}
\email{otani_mayu@cyberagent.co.jp}
\affiliation{%
  \institution{CyberAgent}
  \city{Shibuya-ku}
  \state{Tokyo}
  \country{Japan}
}

\author{Edgar Simo-Serra}
\email{ess@waseda.jp}
\affiliation{%
  \institution{Waseda University}
  \city{Shinjuku-ku}
  \state{Tokyo}
  \country{Japan}
}

\author{Kota Yamaguchi}
\email{yamaguchi_kota@cyberagent.co.jp}
\affiliation{%
  \institution{CyberAgent}
  \city{Shibuya-ku}
  \state{Tokyo}
  \country{Japan}
}

\renewcommand{\shortauthors}{Kotaro Kikuchi et al.}

\begin{abstract}
We introduce MarkupDM, a multimodal markup document model that represents graphic design as an interleaved multimodal document consisting of both markup language and images. Unlike existing holistic approaches that rely on an element-by-attribute grid representation, our representation accommodates variable-length elements, type-dependent attributes, and text content. Inspired by fill-in-the-middle training in code generation, we train the model to complete the missing part of a design document from its surrounding context, allowing it to treat various design tasks in a unified manner. Our model also supports image generation by predicting discrete image tokens through a specialized tokenizer with support for image transparency. We evaluate MarkupDM on three tasks, attribute value, image, and text completion, and demonstrate that it can produce plausible designs consistent with the given context. To further illustrate the flexibility of our approach, we evaluate our approach on a new instruction-guided design completion task where our instruction-tuned MarkupDM compares favorably to state-of-the-art image editing models, especially in textual completion. These findings suggest that multimodal language models with our document representation can serve as a versatile foundation for broad design automation.\footnote{Project page: \url{https://cyberagentailab.github.io/MarkupDM/}}
\end{abstract}


\begin{teaserfigure}
  \centering
  \includegraphics[width=.94\textwidth]{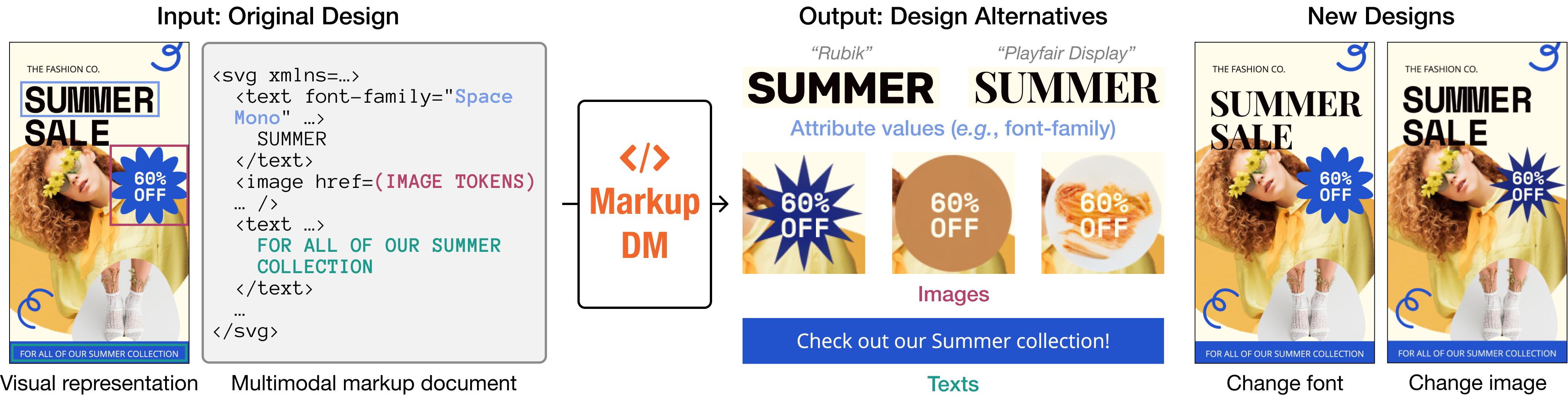}
  \caption{We present a multimodal markup document model (MarkupDM) for graphic design documents. Our model can generate alternative designs by inferring target spans, such as attribute values, images with transparency, and text, from the surrounding context.\todo{Add instruction-guided completion.}}
  \Description{The figure illustrates MarkupDM, a model that generates alternative graphic designs from a markup-based input. It shows an original “SUMMER SALE” poster, its corresponding SVG-like markup code, and the model’s output: design variations created by modifying font styles, images, and promotional text. The final examples demonstrate design updates through font and image changes, while preserving layout and structure.}
  \label{fig:teaser}
\end{teaserfigure}

\maketitle

\section{Introduction}

Graphic design is a visual medium for communicating information and ideas by organizing text, images, and other elements in an aesthetically pleasing way. It is critical in numerous applications, such as websites, advertisements, and printed materials, but creating high-quality designs typically requires specialized expertise and substantial time. Several studies employ machine learning techniques to automate design-related tasks, including layout generation~\cite{InoueLayoutDM2023,HuiUniLayout2023,TangLayoutNUWA2024,seol2024posterllama,LinLayoutPrompter2024,HoritaRetrieval2024,ShabaniVLC2024}, colorization~\cite{QiuColor2023,QiuMultimodal2023,KikuchiGenerative2023}, and typography stylization~\cite{ZhaoModeling2018,ShimodaDiverse2024}. Beyond individual tasks, there have also been holistic modeling approaches for multiple design tasks by formulating graphic design as a grid representation of heterogeneous attributes (element type, position, size, and font information) for each element and then performing generation or completion tasks over this representation~\cite{YamaguchiCanvasVAE2021, InoueFlexible2023}. Although these methods open the door to flexible foundational models for graphic design, they rely on a predefined grid structure and are inefficient for dealing with variable element lengths and type-dependent attributes.

To allow for more flexible application, we represent a graphic design as an interleaved multimodal document composed of markup language and images and then model it using multimodal large language models (LLMs). This resulting formulation is more human-readable and naturally accommodates variable-length elements, type-dependent attributes, and text content. Moreover, by employing the fill-in-the-middle training~\cite{AghajanyanCM32022,BavarianEfficient2022}, we can represent various design tasks in a unified manner by completing the missing part in a document from the surrounding context. We train our model, which we call the \emph{Multimodal Markup Document Model (MarkupDM)}, on 19K graphic design templates. Our model converts image content into discrete tokens using a specialized image tokenizer designed to handle images with transparency in various sizes, allowing it to recognize and generate partial images that compose the overall design. We evaluate MarkupDM on three design completion tasks: generating missing attribute values, images, and text in graphic design templates. Results show that MarkupDM can produce plausible designs consistent with the given context, enabling exploration of various design alternatives (\cref{fig:teaser}).

\looseness=-1
To further demonstrate the extensibility of our approach, we define a new task called instruction-guided graphic design completion, where the model completes a design based on a given instruction. This setup not only reflects the user's intent but also allows an emerging LLM agent~\cite{AgentSurvey} to control the design process, making it more adaptable to specific objectives or creative requirements. To this end, we extend the commonly used Crello dataset~\cite{YamaguchiCanvasVAE2021} to include 125K triplets of instructions, partial designs, and completed designs, resulting in the \emph{Crello-Instruct dataset}. We then fine-tune MarkupDM on this dataset to adapt it to the instruction-guided task. Compared with state-of-the-art image editing models, our model demonstrates favorable performance on this task, particularly in textual completion. Our contributions are as follows:
\begin{itemize}
  \item We formulate graphic design as an interleaved multimodal document consisting of markup language and images.
  \item We propose MarkupDM, a multimodal model that can generate both markup language and images, supported by a tailored image tokenizer capable of encoding  variable-sized images with transparency into discrete tokens.
  \item We extend MarkupDM to an instruction-guided completion task by introducing the Crello-Instruct dataset, which comprises instruction-partial design-completed design triplets.
  \item We show empirically that both MarkupDM and its instruction-tuned variant can successfully complete graphic design documents, demonstrating advantages over existing methods.
\end{itemize}

\section{Related Work}
We first discuss existing approaches to \emph{graphic design generation and completion}, covering both task-specific and holistic modeling methods. We then review recent advances in \emph{multimodal large language models} that can recognize and generate images. Finally, we examine \emph{instruction-guided image editing} methods, clarifying how our structured editing differs from purely image-based approaches.

\subsection{Graphic Design Generation and Completion}
\label{sec:related_graphic_design}
Researchers have long studied computational support for graphic design tasks such as layout generation~\cite{InoueLayoutDM2023,NguyenDiverse2021,ChaiTwoStage2023,ShabaniVLC2024,KikuchiConstrained2021,LinAutoPoster2023}, colorization~\cite{QiuMultimodal2023,KikuchiGenerative2023}, typography stylization~\cite{ZhaoModeling2018,ShimodaDiverse2024}, and general stylization~\cite{ShaoWebRPG2024}. Several studies share a common goal of inferring missing parts or alternative solutions from the existing context. For example, completing a layout from a partially specified layout is a common subtask in layout generation~\cite{InoueLayoutDM2023}. Zhao~\etal~\cite{ZhaoModeling2018} predict typographic styles in web design from both visual and semantic cues. Shao~\etal~\cite{ShaoWebRPG2024} introduce a generative model for web page styling. Qiu~\etal~\cite{QiuMultimodal2023} propose a masked prediction approach for recoloring design documents based on color palette representations.

Different from the task-specific approaches, some studies aim to model entire design documents. CanvasVAE~\cite{YamaguchiCanvasVAE2021} is a variational autoencoder that generates heterogeneous attributes (type, position, size, and image content) for each element in a graphic design document. FlexDM~\cite{InoueFlexible2023} adopts a masked prediction strategy to capture relationships among elements and their attributes. Both methods estimate feature representations for images and text and then retrieve similar ones from a dataset. These methods, however, rely on a predefined element-by-attribute grid representation, which can be inefficient for variable-length elements and type-dependent attributes. There is also growing interest in generating stylized text over generated raster images~\cite{JiaCOLE2024,InoueOpenCOLE2024,WangPrompt2Poster2024,ChenPOSTA2025}, focusing on producing high-quality overall designs.

Recent work has also applied large language models (LLMs) to design tasks~\cite{TangLayoutNUWA2024,seol2024posterllama,LinLayoutPrompter2024,LinParse2023,LinElements2024}. Lin~\etal~\cite{LinParse2023} translate a text description into an intermediate representation to guide the subsequent layout generation. LayoutNUWA~\cite{TangLayoutNUWA2024} formulates layout generation as a code generation task and leverages LLM knowledge to generate layout code. LaDeCo~\cite{LinElements2024} uses multimodal LLMs to automatically place visual and textual elements in a layered manner.

Inspired by these studies, we propose a novel approach to holistic modeling by representing graphic design as an interleaved multimodal document. Unlike the grid-based methods~\cite{InoueFlexible2023,YamaguchiCanvasVAE2021}, our representation naturally accommodates variable-length elements, type-dependent attributes, and text content. We train a multimodal LLM on this document representation and support both text and image generation, in contrast to methods that assume images and text are provided~\cite{LinElements2024} or retrieval-based methods~\cite{InoueFlexible2023,YamaguchiCanvasVAE2021}.

\subsection{Multimodal Large Language Models}
\looseness=-1
The recent success of large language models (LLMs) has led to the development of multimodal LLMs that can recognize and generate images~\cite{YinSurvey2024}. Some approaches, such as DreamLLM~\cite{DongDreamLLM2024}, connect an LLM to an off-the-shelf pre-trained image encoder like CLIP~\cite{RadfordLearning2021} and a decoder such as Stable Diffusion~\cite{RombachHighresolution2022}. However, these image encoders and decoders are not suitable for graphic design tasks because they do not support images with transparency. They also require large-scale image-text datasets, which are difficult to collect in the graphic design domain, where textual descriptions often fail to capture the fine details of images, especially for decorative elements.

Another line of work in multimodal LLMs represents images as discrete tokens~\cite{AghajanyanCM32022,ChameleonTeamChameleon2024,ChernANOLE2024} using a pre-trained image tokenizer like VQGAN~\cite{EsserTaming2021}. Publicly available tokenizers often do not support transparency, but they only require image data rather than large image-text datasets. We adopt this token-based approach and adapt it to handle images with transparency in graphic design.
Furthermore, inspired by LLMs developed for code generation, we use a fill-in-the-middle training objective~\cite{AghajanyanCM32022,BavarianEfficient2022} for our multimodal LLM. This objective enables the model to learn how to complete missing parts of a design from the surrounding context, serving as a flexible foundation for graphic design completion.

\subsection{Instruction-guided Image Editing}
Recent advances in image generation models have led to more practical applications of instruction-guided image editing. InstructPix2Pix~\cite{BrooksInstructpix2pix2023} is a pioneering work in this field. The authors create a dataset by starting with manually created editing examples and then scaling them up using an off-the-shelf large language model and image generation model. MGIE~\cite{FuGuiding2024} augments brief instructions with additional context derived from the embedded knowledge of pre-trained multimodal LLMs. HQ-Edit~\cite{HuiHQEdit2025} enhances dataset quality through a tailored data creation pipeline that leverages advanced foundation models. More recently, proprietary models such as Gemini 2.0 Flash Experimental~\cite{Gemini} and OpenAI's 4o Image Generation~\cite{4oImageGen} have demonstrated impressive performance on image editing tasks. Concurrently, IDEA-Bench~\cite{LiangIDEABench2024} proposes a comprehensive benchmark of professional design tasks, including image retouching and text insertion.

In contrast with the image-based approaches described above, we focus on instruction-guided editing within structured multimodal documents. This approach can improve the preservation of the original content while providing a more interpretable editing process. We build a new dataset specifically for this task and validate our model's performance with it.

\section{Method}
\begin{figure}[t]
  \centering
  \includegraphics[width=\linewidth]{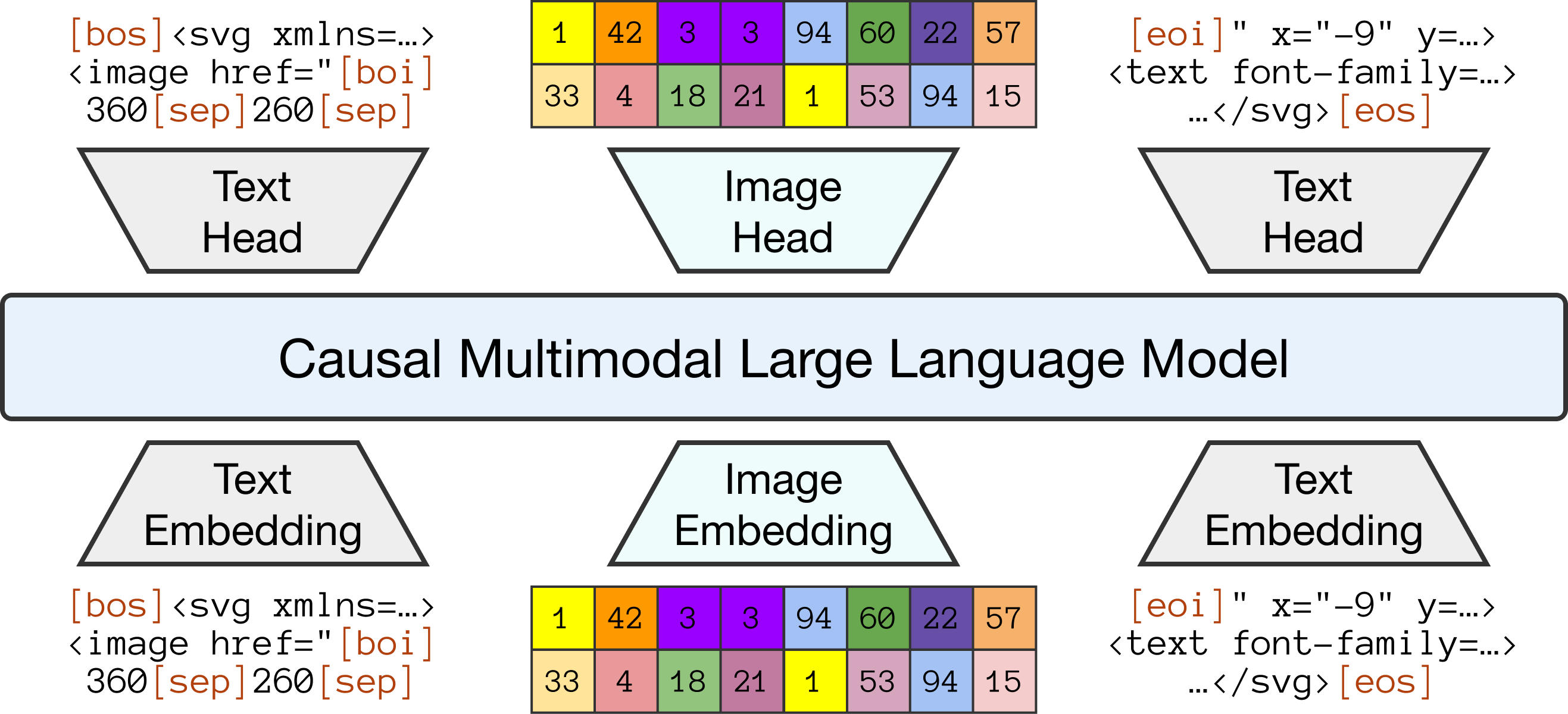}
  \caption{Our MarkupDM is based on causal multimodal LLM, with separate embedding layers and prediction heads dedicated to images and text tokens.}
  \label{fig:overview_mllm}
  \Description{This figure illustrates the architecture of MarkupDM, a causal multimodal large language model. It uses separate embedding layers and prediction heads for text and image tokens within markup documents. SVG markup with embedded image tokens is first split into text and image embeddings, which are processed through a shared LLM core. The output is decoded by modality-specific heads for generating text and image outputs.}
\end{figure}

\begin{figure}
  \centering
  \includegraphics[width=\linewidth]{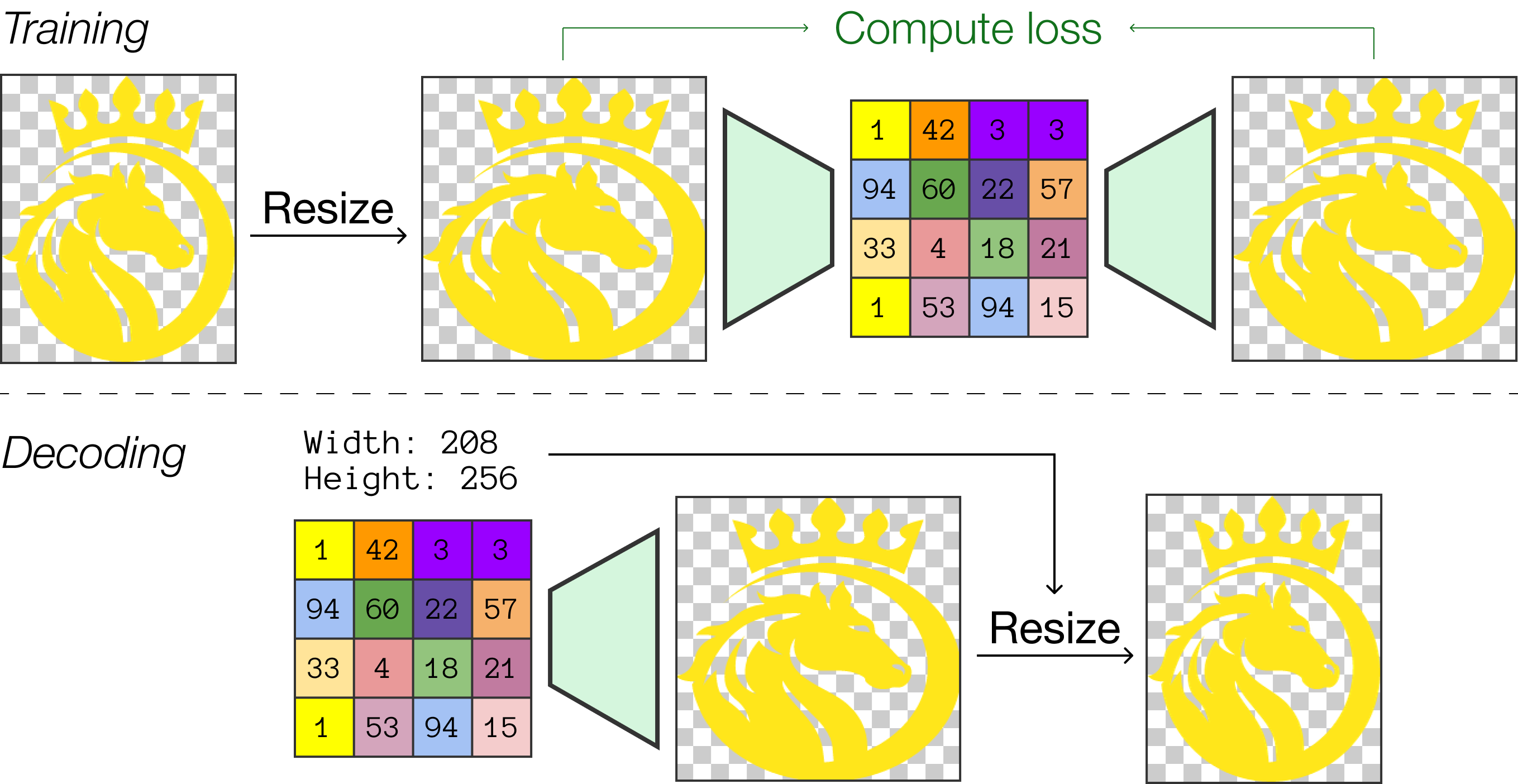}
  \caption{Our image tokenizer is trained by reconstructing images resized to a fixed size. When decoding, the image size is given in addition to the image tokens.}
  \label{fig:overview_autoencoder}
  \Description{This figure explains the training and decoding processes of the image tokenizer. During training, input images are resized to a fixed resolution and reconstructed from image tokens to compute the loss. In decoding, the image is generated from tokens along with explicit width and height information, then resized to the original dimensions.}
\end{figure}

We begin by describing our multimodal document representation, then introduce our proposed MarkupDM model. Finally, we present our specialized image tokenizer, which supports images with transparency commonly used in graphic design. We illustrate an overview of our method in \cref{fig:overview_mllm,fig:overview_autoencoder}.

\subsection{Document Representation} \label{sec:our_representation}
We represent graphic design as a multimodal markup document based on the SVG format\footnote{\url{https://www.w3.org/TR/SVG11/}}, which naturally supports variable-length elements, type-dependent attributes, and text content. Unlike standard SVG, we replace image content with discrete image tokens generated by our image tokenizer (described later in \cref{sec:image_tokenizer}). We show an example of the markup document representation in the following:
\begin{tcolorbox}[
    colframe=gray!90,
    title=Multimodal markup document,
    fonttitle=\small,
    fontupper=\small,
    boxrule=1pt, %
    left=2.5pt, %
    right=2.5pt, %
    top=2.5pt, %
    bottom=2.5pt %
]
\footnotesize
\begin{Verbatim}[commandchars=\\\{\}]
\textcolor{spec}{[bos]}<svg xmlns="http://www.w3.org/2000/svg" viewBox="0 0 419 298 
 "width="419" height="298">
  <image href="\textcolor{spec}{[boi]}360\textcolor{spec}{[sep]}260\textcolor{spec}{[sep]}\textcolor{img}{[img:1][img:42][img:3][img:94}
  \textcolor{img}{]}...\textcolor{spec}{[eoi]}" x="-9" y="-9" width="436" height="315"/>
  <text font-family="Montserrat" font-size="30" font-weight="bold
  " fill="rgba(255, 255, 255, 1)" x="32" y="81">FAMILY</text>
...</svg>\textcolor{spec}{[eos]}
\end{Verbatim}
\end{tcolorbox}

The image content, \ie, the value of {\small\verb|href|} attribute in the {\small\verb|<image>|} tag, starts with the special token {\small\verb|[boi]|} and ends with {\small\verb|[eoi]|}.
The inside of these is separated by the special token {\small\verb|[sep]|}, and each represents the width, height, and image tokens such as {\small\verb|[img:1]|} obtained by our image tokenizer.
This image representation is similar to the previous work on a multimodal LLM for simplified HTML documents~\cite{AghajanyanCM32022}, but differs in that the image size is also described as text and included in the target of generation.

\subsection{Multimodal Markup Document Model} \label{sec:MarkupDM}
To incorporate the image representation described in \Cref{sec:our_representation}, we build the multimodal markup document model (MarkupDM) by applying two extensions to the base LLM. 
First, we extend the vocabulary of the base LLM to include the additional special tokens, such as {\small\verb|[boi]|}.
Second, we add new modules dedicated to the image tokens, such as {\small\verb|[img:1]|}, the embedding module, and the prediction head.
In the embedding module, we first embed the image tokens via the frozen lookup table in our image decoder (\cref{sec:image_tokenizer}). We then concatenate them with the positional encodings~\cite{TancikFourier2020} and project them to the same dimension as the text embeddings.
The prediction head for image tokens is similar to the one for text tokens, but uses a different set of parameters and vocabulary, \ie, the codebook size in image tokenization.

We train our model based on the next token prediction in our sequences to which we randomly apply the fill-in-the-middle transformation~\cite{BavarianEfficient2022,AghajanyanCM32022}, allowing the model to predict the missing middle part from the prefix and suffix parts.
During inference, our model must identify the modality of the next token due to the different prediction heads. To determine which modality to generate, we explicitly track whether the model is currently in the process of generating image tokens based on the generated text so far.

\subsection{Specialized Image Tokenizer} \label{sec:image_tokenizer}
Existing publicly available image tokenizers are typically designed for RGB images and thus do not support transparency in images, which is common in graphic design. To address this limitation, we develop a new image tokenizer by training an image autoencoder that encodes transparent images of varying sizes into discrete token maps at a $1/f$ resolution and decodes them back into the original images.
While it is straightforward to vary the token size according to the image size, we found in preliminary experiments that this makes it difficult to train the markup language model at a later stage. 
Instead, we take a simple but effective approach of resizing the input image into a fixed square size.
We follow the previous studies~\cite{EsserTaming2021,RombachHighresolution2022} and take the same network architecture and training objectives for our autoencoder, with the only difference related to the alpha channel, \ie, transparency.
We set the number of input/output channels to four and consider L1 reconstruction loss for all channels.
When calculating the loss based on RGB-based external models, \eg, the perceptual loss~\cite{ZhangUnreasonable2018}, we convert generated RGBA images to RGB images by alpha compositing on a white background.
We initialize our model with the weights of a pre-trained RGB image tokenizer. For the alpha channel weights, we use the mean values of the corresponding RGB weights.

\section{Crello-Instruct Dataset} \label{sec:dataset}
In addition to the document completion tasks and to showcase the extensibility of our approach, we introduce a new task called instruction-guided graphic design completion, which requires the model to complete a design based on a provided instruction. To create the benchmark dataset for this task, we extend the commonly used Crello dataset~\cite{YamaguchiCanvasVAE2021} to support instruction-guided completion. We refer to the resulting dataset as the \emph{Crello-Instruct dataset}.  A design template in the Crello dataset includes multimodal elements such as text, images, and other visual elements. We remove one of the elements to create a partial design, then use the specialized renderer\footnote{\url{https://github.com/CyberAgentAILab/cr-renderer}} to generate rendered images of both the partial and original designs. Then, we feed the partial and original designs into the Qwen2.5-VL-7B-Instruct model~\cite{qwen25vl} and ask it to generate an instruction to recreate the original design based on the partial design. Because the resulting instructions are often noisy, we use GPT-4o mini~\cite{GPT4oMini} to rate the quality of each triplet (instruction, partial design, and completed design) and filter out lower-quality samples. We then use the filtered dataset to train and evaluate our instruction-tuned model.

Additionally, we generate a caption for each non-textual element in the dataset with Qwen2.5-VL-7B-Instruct~\cite{qwen25vl} to help the model understand the image content. In our document representation, we add an extra {\small\verb|caption|} attribute in the {\small\verb|<image>|} tag, placing it before the {\small\verb|href|} attribute, so that the model predicts the caption first and then the actual image tokens~\cite{AghajanyanCM32022}. We provide examples of instructions and captions in \cref{fig:dataset_example}. The caption examples highlight the unique challenges of this dataset, which contains both semantically describable elements and abstract decorative ones, and the later often have noisy captions. Further details can be found in the supplementary material.

\setlength{\tabcolsep}{2pt}
\newcommand{\frfig}[2]{\frame{\includegraphics[width=#1\linewidth]{figures/#2}}}
\begin{figure}[t]
    \centering
    \begin{subfigure}[b]{\linewidth}
      \centering
      \begin{tabular}{C{25mm}C{25mm}C{25mm}}
        \frfig{1}{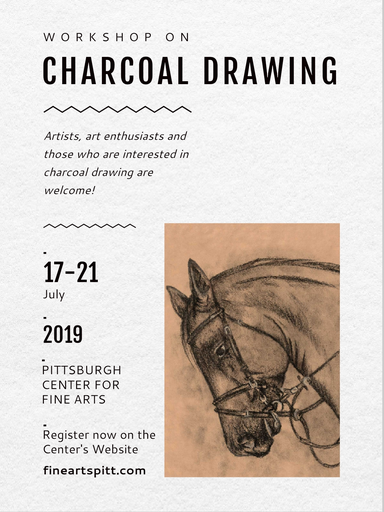} &
        \frfig{1}{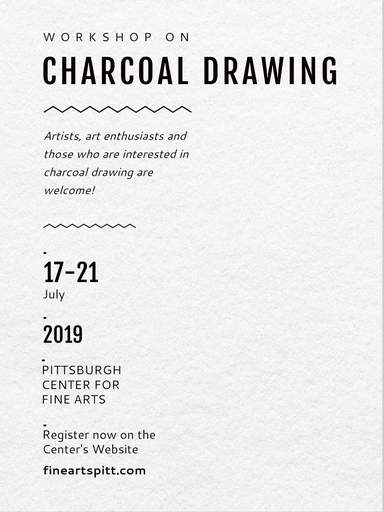} &
        \frfig{1}{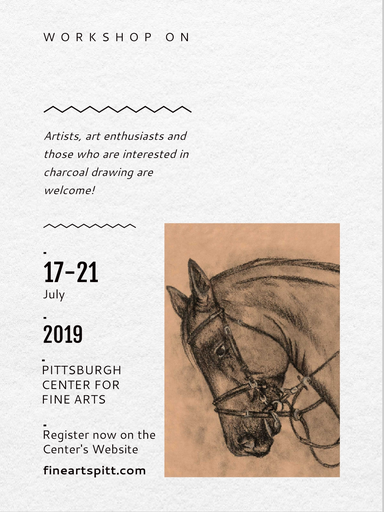} \\
        {\small Completed design} &
        {\small Partial 1: \textit{Add a charcoal drawing of a horse's head in the bottom right corner of the image.}} &
        {\small Partial 2: \textit{Replace ``WORKSHOP ON'' with ``WORKSHOP ON CHARCOAL DRAWING''.}} \\
      \end{tabular}
      \caption{Completed design and two partial designs with instructions.}
      \Description{This figure shows a completed design and two partial variants accompanied by editing instructions. The completed poster promotes a charcoal drawing workshop. Partial 1 omits the horse image and includes an instruction to add a charcoal horse drawing in the bottom right. Partial 2 lacks the full title and includes an instruction to replace “WORKSHOP ON” with “WORKSHOP ON CHARCOAL DRAWING”.}
    \end{subfigure}
    \vskip\baselineskip
    \begin{subfigure}[b]{\linewidth}
      \centering
      \begin{tabular}{C{17mm}L{58mm}}
        \frfig{1}{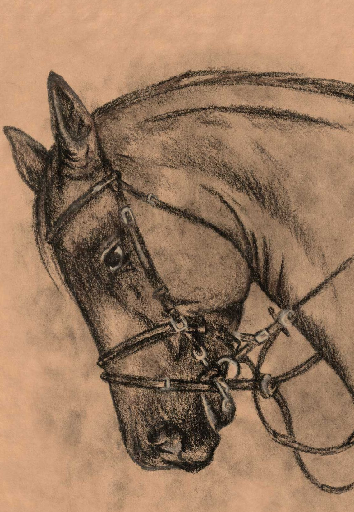} &
        {\small \textit{A detailed pencil sketch of a horse's head and part of its neck. The horse is wearing a bridle with reins, ...}}

        \begin{minipage}[b]{\linewidth}
          \raggedleft
          \begin{minipage}[b]{.95\linewidth}
            \vspace{4mm}
            \raggedleft
            \frfig{1}{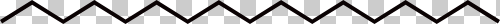} \\
            {\small \textit{A zigzag pattern consisting of alternating straight and curved segments, ...}}
          \end{minipage}
        \end{minipage}
      \end{tabular}
      \caption{Image elements with captions.}
      \Description{This figure presents two individual image elements with corresponding captions. The first is a pencil sketch of a horse’s head and neck, featuring a bridle with reins. The second is a zigzag line pattern composed of alternating straight and curved segments, illustrating a decorative visual motif.}
    \end{subfigure}
    \caption{Examples of our Crello-Instruct dataset.}
    \label{fig:dataset_example}
\end{figure}
\setlength{\tabcolsep}{6pt}

\section{Experiments}

We begin by evaluating our image tokenizer on an image reconstruction task. Next, we assess our multimodal markup language models on various graphic design completion tasks. Finally, we evaluate our instruction-tuned models on the instruction-guided completion task.

\subsection{Image Reconstruction}

\subsubsection{Setup} \label{sec:image_tokenizer_setup}
We use an internal dataset of graphic design templates, which is similar to the Crello dataset~\cite{YamaguchiCanvasVAE2021}.
Each template consists of an ordered set of elements, and each element is associated with an element category, geometric attributes, and design attributes. The template also includes global attributes such as canvas size. We use 800,000 RGBA images of non-textual elements from these design templates for training and 133,267 images from different templates for evaluation.

We finetune a baseline RGB tokenizer for 100,000 steps, following the techniques explained in \cref{sec:image_tokenizer}, to adapt it to RGBA images. For the baseline tokenizer, we adopt the one from the Latent Diffusion Model (LDM-VQ)~\cite{RombachHighresolution2022} trained on the OpenImages dataset~\cite{KuznetsovaOpen2020}, which is primarily composed of photographs. Specifically, we use the tokenizer with the scaling factor $f\!=\!16$ and the codebook size $Z\!=\!16{,}384$, balancing reconstruction quality and the resulting token length. For further analysis, we finetune the tokenizer solely on RGB images without additional techniques, referred to as \emph{Ours-RGB}. As an additional baseline without the specialized tokenizer, we convert RGB images into RGBA using an off-the-shelf background removal tool, Rembg~\cite{GatisRembg2020} with IS-Net~\cite{QinHighly2022}.

We evaluate the tokenizers using mean squared error (MSE) for both the RGB and alpha channels, as well as reconstruction Fr\'echet Inception Distance (rFID) for RGB images, which measures the distance between the feature distributions of the original and reconstructed images. For RGB-based metrics, we convert the RGBA images generated by our tokenizer to RGB by alpha compositing them onto a white background.

\subsubsection{Results}
\begin{figure}[t]
    \centering
    \begin{tabular}{ccc}
        {\small Ours-RGB + Rembg~\cite{GatisRembg2020}} & {\small Ours} & {\small Original} \\
        \frfig{.333}{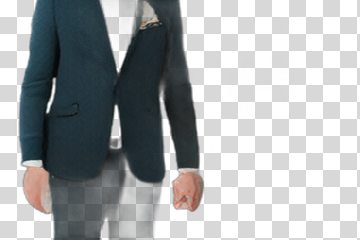} &
        \frfig{.333}{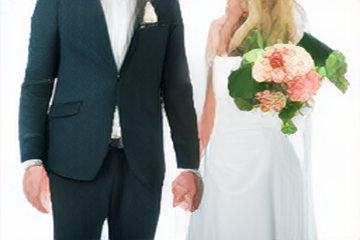} &
        \frfig{.333}{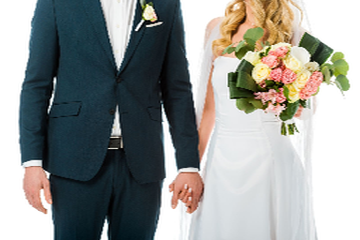} \\[-2.8pt]
        \frfig{.1726}{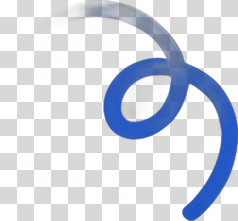}\frfig{.1604}{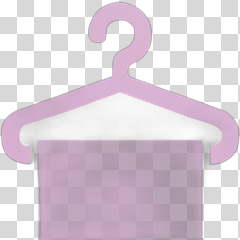} &
        \frfig{.1726}{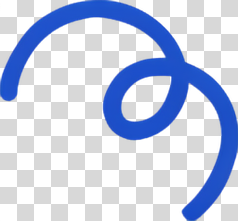}\frfig{.1604}{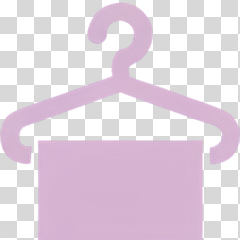} &
        \frfig{.1726}{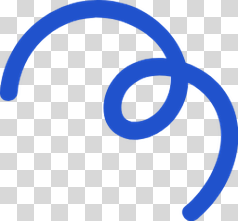}\frfig{.1604}{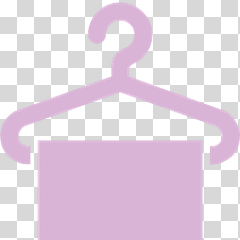} 
    \end{tabular}
    \caption{Image reconstruction results.}
    \label{fig:image_reconstruction}
    \Description{This figure contains two rows of three images each. The columns are labeled (from left to right) “Ours-RGB + Rembg,” “Ours,” and “Original.” In the top row, the images show a wedding couple; in the bottom row, the images show a stylized coat hanger and swirl graphic. Compared with the original (right), “Ours-RGB + Rembg” (left) exhibits visual artifacts from background removal, often leaving traces around the subjects or removing portions of the foreground. “Ours” (middle) more accurately reconstructs the RGBA images, preserving foreground objects and transparent areas.}
\end{figure}

We show a quantitative comparison of image reconstruction in \cref{tab:image_reconstruction}. Both of our tokenizers outperform the baseline in terms of RGB-based metrics thanks to their fine-tuning on images from the same domain. We also show qualitative comparisons in \cref{fig:image_reconstruction}. As illustrated, the general background removal used for RGB-based reconstructions often fails, removing foreground objects either too aggressively or insufficiently. In contrast, our tokenizer successfully reconstructs RGBA images by leveraging the alpha information embedded in the discrete tokens.

\begin{table}
  \centering
  \caption{Quantitative comparison of image reconstruction for each tokenizer. The dagger symbol ($\dagger$) indicates the score computed by setting the alpha value of every pixel to 1.0.}
  \label{tab:image_reconstruction}
  \begin{tabular}{lccc}
  \toprule
  & \multicolumn{2}{c}{MSE $\downarrow$} & rFID $\downarrow$ \\
  & RGB {\scriptsize ($\times\textrm{10}^\textrm{-3}$) }
  & Alpha {\scriptsize ($\times\textrm{10}^\textrm{-1}$) }
  & RGB \\
  \midrule
  LDM-VQ~\cite{RombachHighresolution2022} & 2.42 & 3.75$^\dagger$ & 6.34 \\
  Ours-RGB & \textbf{1.50} & 3.75$^\dagger$ & \textbf{1.65} \\
  Ours & \underline{1.86} & \textbf{0.03} & \underline{4.96} \\
  \bottomrule
  \end{tabular}
\end{table}

\setlength{\tabcolsep}{0pt}
\setlength{\tabcolsep}{6pt}

\subsection{Graphic Design Completion}

\subsubsection{Setup} \label{sec:markupdm_setup}
We use the Crello dataset~\cite{YamaguchiCanvasVAE2021} (version 5.0.0), comprising 19,372 templates for training, 1,823 for validation, and 2,107 for testing. We then convert these templates into SVG format. During the conversion, we represent text elements with {\small\verb|<text>|} tags and other elements with {\small\verb|<image>|} tags. We omit attributes when they have default values. Also, because SVG does not support multi-line text within a single element, we split any text element into multiple elements whenever a new line appears.

We train our MarkupDM with the fill-in-the-middle (FIM) objective~\cite{BavarianEfficient2022,AghajanyanCM32022}, which predicts a randomly selected middle span based on the prefix and suffix. In this setup, MarkupDM can infer the missing span from its preceding and following context. To demonstrate its effectiveness, we evaluate three tasks: attribute value completion, image completion, and text completion. Attribute value completion is represented as {\small\verb|<text x="[MASK]" ...>|}, where {\small\verb|[MASK]|} indicates the span to be filled. Image completion is represented as {\small\verb|<image href="[MASK]" .../>|}, and text completion is represented as {\small\verb|<text ...>[MASK]</text>|}. For attribute value completion, we focus on six attribute types: {\small\verb|x|}, {\small\verb|y|}, {\small\verb|width|}, {\small\verb|height|}, {\small\verb|font-family|}, and {\small\verb|font-size|}. Note that we do not train MarkupDM with task-specific supervision such as specialized FIM patterns; these tasks serve only for post-hoc evaluation.

We evaluate MarkupDM using several base language models, including StarCoderBase~\cite{LiStarCoder2023} with 1B, 3B, and 7B parameters, as well as Qwen2.5-7B~\cite{qwen25} and Qwen2.5-Coder-7B~\cite{qwen25coder}. We specifically select these models because they provide sufficiently long context lengths and employ the FIM objective during their pre-training. Both of the features are essential for our completion tasks where the model must handle multiple textual and visual elements and dynamically insert missing parts. For comparison with the approach of holistic yet grid-based graphic design generation approach (see \Cref{sec:related_graphic_design}), we also train FlexDM~\cite{InoueFlexible2023} on our dataset using random masking patterns, aiming to create similar experimental conditions. Note that during text and image completion tasks, FlexDM retrieves texts or images from the combined train and validation set instead of generating them directly.

We evaluate MarkupDM on between 12,559 and 25,435 target spans from the test templates, selecting the relevant spans for each task. To reduce inference time for image completion, we use 1,386 spans from the first 200 templates. We parse the text generated by MarkupDM and convert it to the same format used by FlexDM. We then compute accuracy over the quantized representation for attribute value completion, and cosine similarity over feature representations for text and image completion. More details are provided in the supplementary material.

\begin{table*}[t]
  \caption{Quantitative comparison for design completion tasks. The reported scores reflect accuracy for attribute values and cosine similarity for text and image completion. ``Font'' denotes the font family, and ``F-Size'' denotes the font size. ``Mean'' indicates the average score of all the completion tasks. FlexDM follows a different formulation than MarkupDM, so its scores are not directly comparable and are provided only for reference.}
  \label{tab:design_completion}
  \begin{tabular}{lc cccccc cc c}
  \toprule
  Model & Base LLM & X $\uparrow$ & Y $\uparrow$ & Width $\uparrow$ & Height $\uparrow$ & Font $\uparrow$ & F-Size $\uparrow$ & Text $\uparrow$ & Image $\uparrow$ & Mean $\uparrow$ \\
  \midrule
  FlexDM~\cite{InoueFlexible2023} & -- & 0.420 & 0.268 & 0.406 & 0.612 & 0.844 & \textbf{0.851} & 0.813 & 0.759 & 0.622 \\ \midrule
   & Qwen2.5-7B & 0.460 & 0.285 & 0.824 & 0.904 & 0.460 & 0.670 & 0.827 & 0.811 & 0.655 \\
   & Qwen2.5-Coder-7B & 0.486 & 0.331 & 0.853 & 0.931 & 0.365 & 0.700 & 0.851 & 0.806 & 0.665 \\
   MarkupDM & StarCoderBase-1B & 0.471 & 0.339 & 0.843 & 0.920 & 0.845 & 0.678 & 0.851 & \underline{0.822} & 0.721 \\
   & StarCoderBase-3B & \underline{0.508} & \underline{0.379} & \underline{0.870} & \underline{0.936} & \underline{0.854} & \underline{0.724} & \underline{0.865} & \textbf{0.823} & \underline{0.745} \\
   & StarCoderBase-7B & \textbf{0.526} & \textbf{0.404} & \textbf{0.882} & \textbf{0.951} & \textbf{0.867} & 0.720 & \textbf{0.874} & 0.817 & \textbf{0.755} \\
   \bottomrule
  \end{tabular}
\end{table*}

\subsubsection{Results for Attribute Value Completion}
We show the quantitative results for attribute values (\emph{X}, \emph{Y}, \emph{Width}, \emph{Height}, \emph{Font}, \emph{F-Size}) in \cref{tab:design_completion}. Note that the scores for FlexDM and MarkupDM are not fully comparable, because they differ in formulation and available contextual cues. For example, MarkupDM can infer element sizes from the image dimensions, whereas FlexDM cannot. Nevertheless, MarkupDM performs well in comparison, indicating that it successfully learns to fill graphic design templates. Among the MarkupDM variants, StarCoderBase-7B achieves the highest accuracy for most attributes. Comparing the results across different parameter sizes of StarCoderBase (1B, 3B, and 7B), we observe that larger models consistently perform better, as expected. Although Qwen2.5-based models also work with our approach, they tend to show lower performance, possibly due to limited exposure to SVG data during pre-training.

\subsubsection{Results for Text Completion}
\begin{figure*}[t]
    \centering
    \includegraphics[width=.94\linewidth]{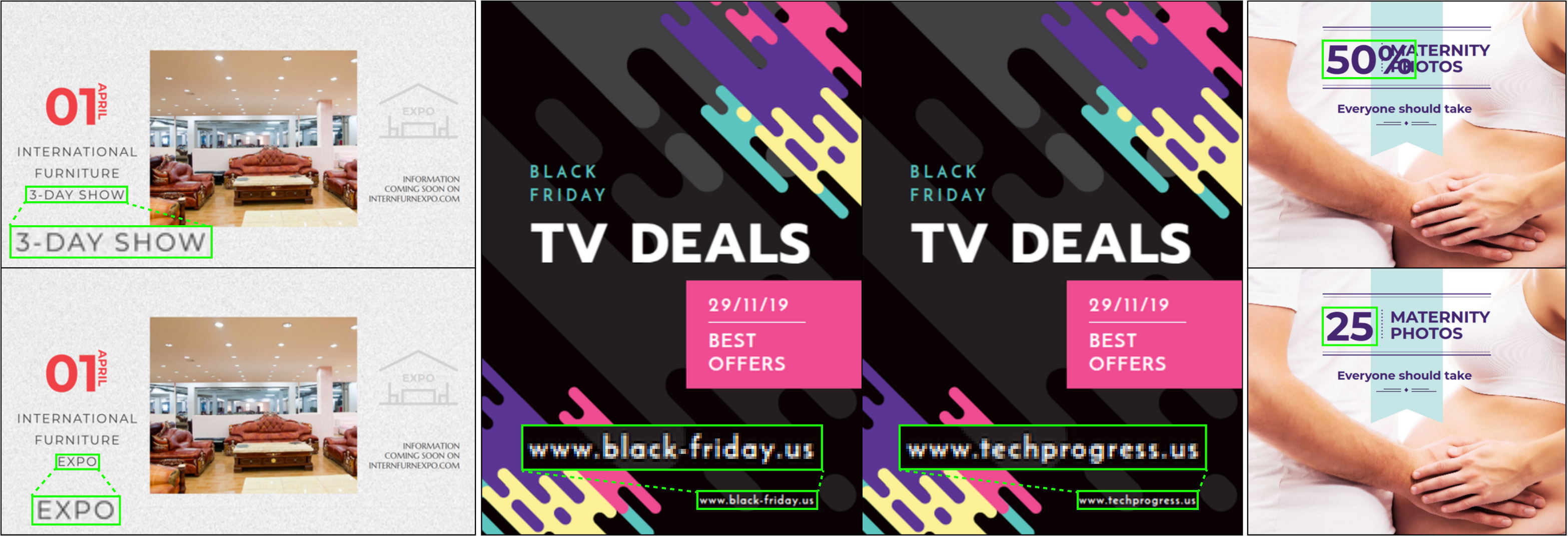}
    \caption{Text completion results. Each pair shows the predicted completion and the original design from left to right or top to bottom. The green boxes indicate the target text and some of them are zoomed in for better visibility.}
    \label{fig:text_completion}
    \Description{This figure shows text completion results, where each pair presents the input design and the predicted completion. Green boxes highlight the target text regions. Most completions accurately match the original design, including titles and website URLs. However, the rightmost example illustrates a failure case: the predicted number "50 \%" collides with the surrounding text, creating a visual overlap that disrupts legibility and layout integrity.}
\end{figure*}

\looseness=-1
We present the quantitative results for text completion in the \emph{Text} column of \cref{tab:design_completion}. We observe that our model outperforms the baseline, and its performance improves as the model size increases. In the left and middle parts of \cref{fig:text_completion}, we show examples where the model successfully generates text that aligns grammatically with preceding or subsequent lines, or that serves a similar role to the ground truth text. Our model sometimes fails due to errors in image understanding or conflicting with other elements visually, \eg, the rightmost example.

\newcommand{\fighh}[2]{
  \frfig{#2}{design_completion/#1_pred.png}&
  \frfig{#2}{design_completion/#1_gt.png}
}
\newcommand{\figvv}[3]{
  \multirow[t]{1}{#2\linewidth}[#3]{
    \frfig{1}{design_completion/#1_pred.png}\\[-1pt]
    \frfig{1}{design_completion/#1_gt.png}
  }
}
\setlength{\tabcolsep}{0pt}
\setlength{\tabcolsep}{6pt}

\newcommand{\figh}[2]{
  \frfig{#2}{design_completion/#1_input.png}&
  \frfig{#2}{design_completion/#1_pred.png}&
  \frfig{#2}{design_completion/#1_gt.png}
}
\newcommand{\figv}[3]{
  \multirow[t]{1}{#2\linewidth}[#3]{
    \frfig{1}{design_completion/#1_input.png}\\[-1pt]
    \frfig{1}{design_completion/#1_pred.png}\\[-1pt]
    \frfig{1}{design_completion/#1_gt.png}
  }
}
\setlength{\tabcolsep}{0pt}
\begin{figure*}[t]
    \centering
    \includegraphics[width=.94\linewidth]{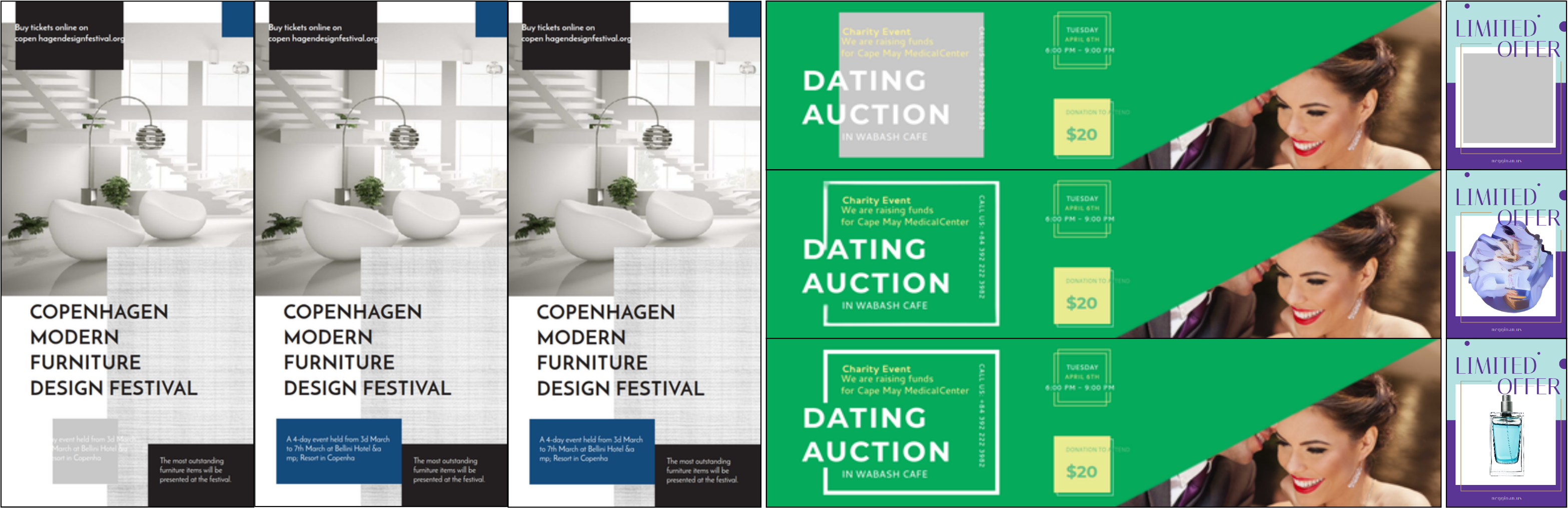}
    \caption{Image completion results. Each triplet shows the input, the predicted completion, and the original design from left to right or top to bottom. The gray squares indicate the target image elements to be completed.}
    \label{fig:image_completion}
    \Description{This figure shows image completion results. Each triplet displays the input with missing regions (gray squares), the model’s predicted completion, and the original design. Examples include furniture advertisements, event posters, and product promotions. The predictions aim to reconstruct visual content such as background textures, white line decor, or product images based on surrounding layout and context. The rightmost shows a failure case: the generated image is incoherent and lacks recognizable content compared to the original design.}
\end{figure*}
\setlength{\tabcolsep}{6pt}

\subsubsection{Results for Image Completion}
\looseness=-1
The quantitative results for image completion in the \emph{Image} column of \Cref{tab:design_completion} also demonstrate improved performance compared to the baseline. Unlike text completion, however, the variation in performance with respect to model size is relatively smaller.
For deeper analysis, we investigate the effect of providing auxiliary caption information, which we introduced in \cref{sec:dataset}. In \cref{tab:ablation_caption}, we observe no performance gain when training the model with captions. However, using ground-truth captions substantially improves image generation performance (the bottom row of \cref{tab:ablation_caption}), suggesting that the model struggles to accurately predict content, possibly due to limited training data. Qualitative results in \cref{fig:image_completion} illustrate that our model can generate simpler design elements, such as underlays or buttons, by leveraging textual content or repetition patterns as hints. Our model has difficulty in producing main objects like the rightmost example or delicate visual harmonization with other elements. For example, in the middle example, the generated decoration slightly conflicts with the text element, highlighting the need for visual feedback.

\begin{table}[t]
  \caption{Image completion results using captions as auxiliary information. The baseline model is MarkupDM with StarCoderBase-7B.}
  \label{tab:ablation_caption}
\begin{adjustbox}{max width=1\columnwidth}
  \begin{tabular}{cccc}
  \toprule
  Train with Caption & Test Input & Completion Target & Image $\uparrow$ \\
  \midrule
  -- & Context & Image & 0.817 \\
  \ding{51} & Context & Caption + Image & 0.815 \\
  \ding{51} & Context + Caption & Image & 0.857 \\
  \bottomrule
  \end{tabular}
\end{adjustbox}
\end{table}

\subsection{Instruction-Guided Completion}

\subsubsection{Setup} \label{sec:instruction_setup}
We use the Crello-Instruct dataset as described in \cref{sec:dataset}. Each sample is a triplet composed of an input document with one element missing, an instruction for completing that document, and a target document in which the missing element is filled in. The dataset includes 103,917 samples for training, 9,839 for validation, and 11,350 for testing.

\looseness=-1
We fine-tune the best variant of MarkupDM, \ie, the one that uses StarCoderBase-7B as its base LLM, on this dataset, referring to the resulting model as \emph{Instruct-MarkupDM}. For our baselines, we select two image editing methods: HQ-Edit~\cite{HuiHQEdit2025} and Gemini 2.0 Flash Experimental (Gemini 2.0 FE)~\cite{Gemini}. HQ-Edit is one of the latest open-source image editing models; we use both its original pre-trained model and a version further fine-tuned on our dataset. Gemini 2.0 FE is a proprietary model, which has recently demonstrated strong performance in terms of both image quality and instruction adherence.

We evaluate each model's performance using four pixel-based metrics: MSE$_\mathrm{GT}$, MSE$_\mathrm{Edit}$, Alignment~\cite{HuiHQEdit2025}, and Coherence~\cite{HuiHQEdit2025}. MSE$_\mathrm{GT}$ measures the pixel-wise difference between the predicted image and the ground truth image, while MSE$_\mathrm{Edit}$ measures the difference between the input and the predicted image. A lower MSE$_\mathrm{Edit}$ than the ground-truth score indicates that the model has under-edited the image, whereas a higher MSE$_\mathrm{Edit}$ suggests over-editing or adding irrelevant elements. Therefore, while a lower score indicates better performance for MSE$_\mathrm{GT}$, MSE$_\mathrm{Edit}$ is considered better when its score is closer to the ground truth score. Alignment and Coherence~\cite{HuiHQEdit2025} are both GPT-based evaluations: Alignment measures the degree to which the edited image satisfies the instruction in the context of the input image, and Coherence assesses the overall visual quality of the edited image, independent of the instruction. We employ GPT-4o mini~\cite{GPT4oMini} for both metrics, using the same prompts specified in the previous work~\cite{HuiHQEdit2025}.

\subsubsection{Results} \label{sec:instruction_results}

\begin{figure*}[t]
    \centering
    \begin{tabular}{cccccc}
      Input & HQ-Edit~\cite{HuiHQEdit2025} & HQ-Edit~\cite{HuiHQEdit2025} + FT & Gemini 2.0 FE~\cite{Gemini} & Instruct-MarkupDM & Ground Truth \\
      \frfig{.148}{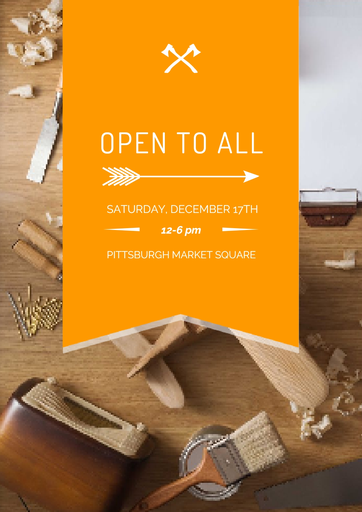} &
      \frfig{.148}{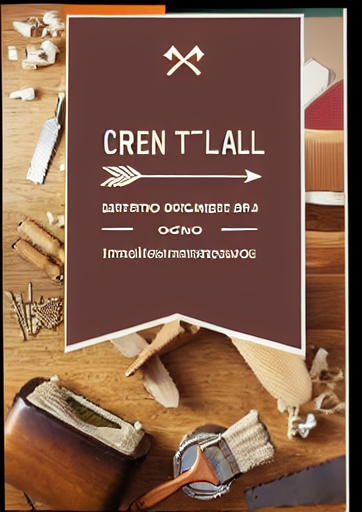} &
      \frfig{.148}{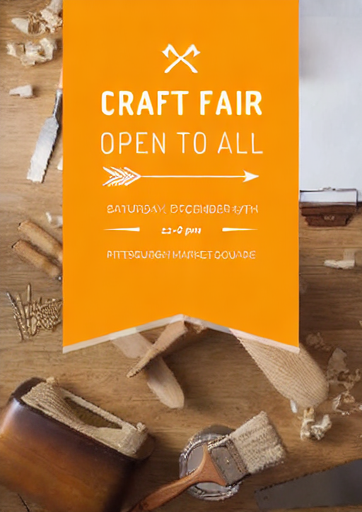} &
      \frfig{.148}{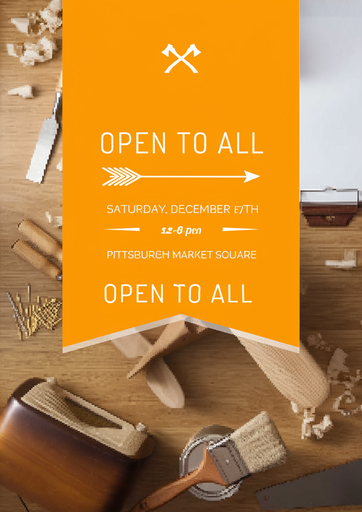} &
      \frfig{.148}{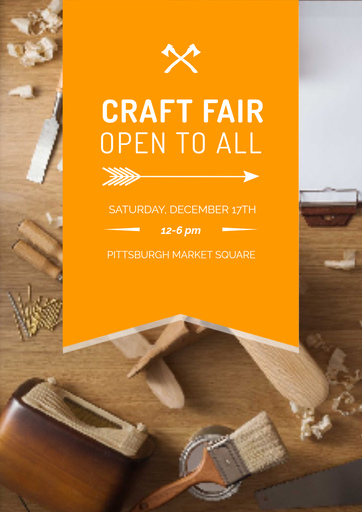} &
      \frfig{.148}{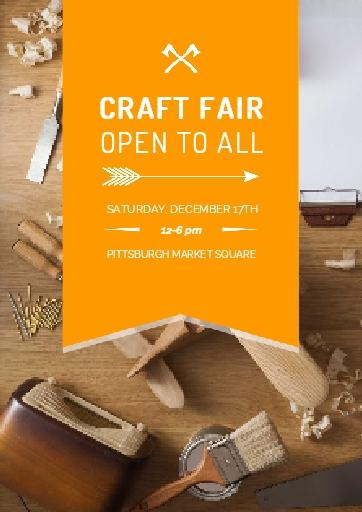} \\
      \multicolumn{6}{c}{\textit{Add ``CRAFT FAIR'' above ``OPEN TO ALL'' in the central orange section of the image.}} \\[6pt]
      \frfig{.148}{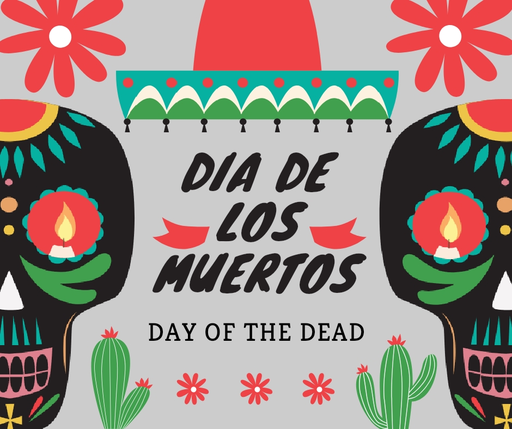} &
      \frfig{.148}{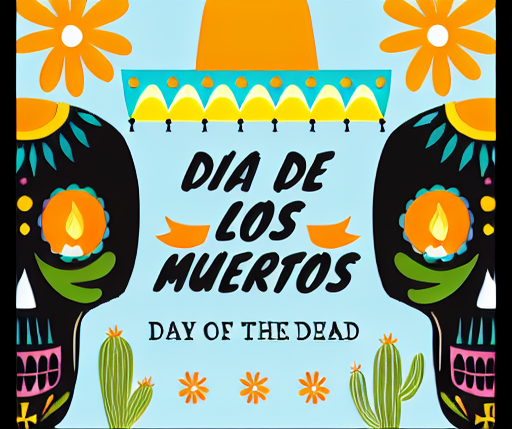} &
      \frfig{.148}{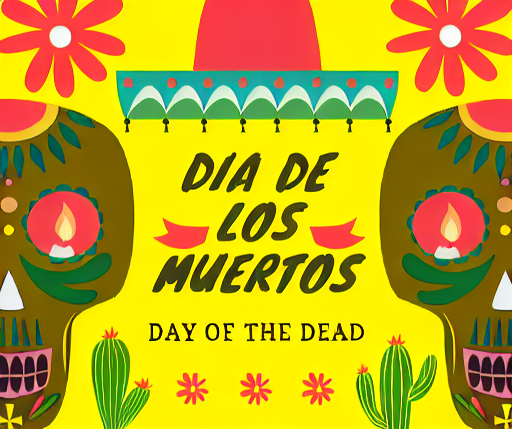} &
      \frfig{.148}{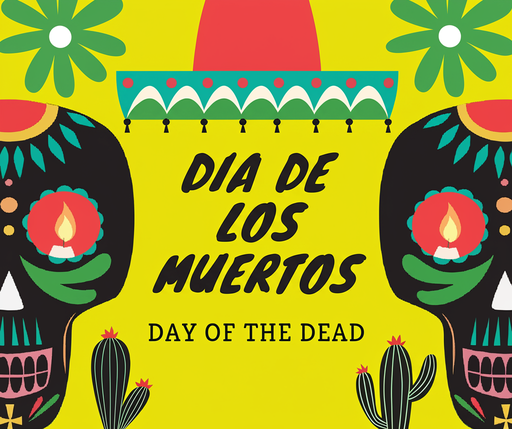} &
      \frfig{.148}{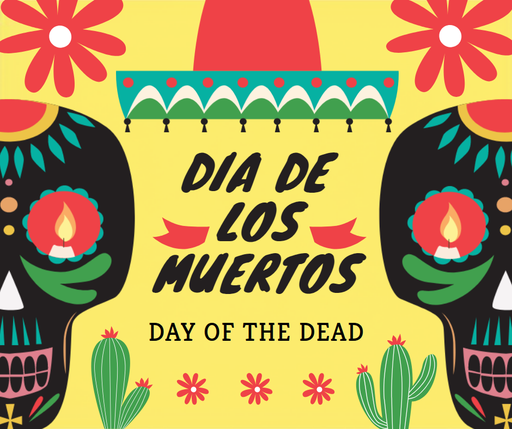} &
      \frfig{.148}{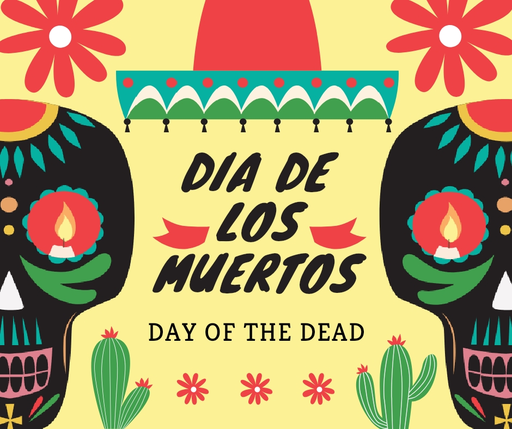} \\
      \multicolumn{6}{c}{\textit{Change the background color of the image to yellow.}}
    \end{tabular}
    \caption{Qualitative comparison for instruction-guided graphic design completion.}
    \label{fig:instructed_completion}
    \Description{This figure compares instruction-guided graphic design completions across several models. The top row shows an input flyer with the instruction to insert “CRAFT FAIR” above “OPEN TO ALL”. The bottom row presents a second instruction to change the background color to yellow in a “DIA DE LOS MUERTOS” design.}
\end{figure*}

We present the quantitative results in \cref{tab:instructed_completion} and the qualitative results in \cref{fig:instructed_completion} for instruction-guided graphic design completion. Among the image-editing methods, HQ-Edit highlights the importance of fine-tuning on our design dataset to bridge the domain gap from general image-editing datasets. By contrast, Gemini 2.0 FE achieves better performance than HQ-Edit even in zero-shot settings, presumably due to its strong instruction-following and image-generation capabilities. However, Gemini 2.0 FE sometimes applies overly aggressive visual edits or incorrect text edits (as shown in the top example in \cref{fig:instructed_completion}), leading to poor MSE scores.

Instruct-MarkupDM achieves the best MSE scores and a higher Coherence score, because it only adds the missing elements rather than altering existing ones, leaving most input designs intact. However, its Alignment score is lower than that of Gemini 2.0 FE, possibly reflecting less robust instruction-following and visual generation capabilities. As \cref{fig:instructed_completion} illustrates, Instruct-MarkupDM generally handles text editing well but struggles with generating complex visual elements beyond simple colored backgrounds.

\setlength{\tabcolsep}{4pt}
\begin{table}
\caption{Quantitative comparison for instruction-guided graphic design completion.}
\label{tab:instructed_completion}
\begin{tabular}{l cccc}
\toprule
Model & MSE$_\mathrm{GT}$ $\downarrow$ & MSE$_\mathrm{Edit}$ & Align. $\uparrow$ & Coher. $\uparrow$ \\
\midrule
HQ-Edit~\cite{HuiHQEdit2025} & 93.9 & 93.5 & 28.6 & 57.4 \\
~+ Finetune & 43.9 & 43.1 & 51.1 & 62.3 \\
Gemini 2.0 FE~\cite{Gemini} & \underline{33.5} & \underline{31.6} & \textbf{72.3} & \textbf{69.4} \\
Instruct-MarkupDM & \textbf{10.0} & \textbf{6.7} & \underline{60.5} & \underline{69.3} \\ \midrule
Ground Truth & 0.0 & 8.2 & 85.2 & 71.8 \\
\bottomrule
\end{tabular}
\end{table}
\setlength{\tabcolsep}{6pt}

\setlength{\tabcolsep}{1.5pt}
\setlength{\tabcolsep}{6pt}

\begin{figure}[t]
    \centering
    {\small \begin{tabular}{ccc}
      & Instruct-MarkupDM$^\ast$ & \\
      Instruct-MarkupDM$^\ast$ & + T2I~\cite{Gemini} & Ground Truth \\
      \frfig{.32}{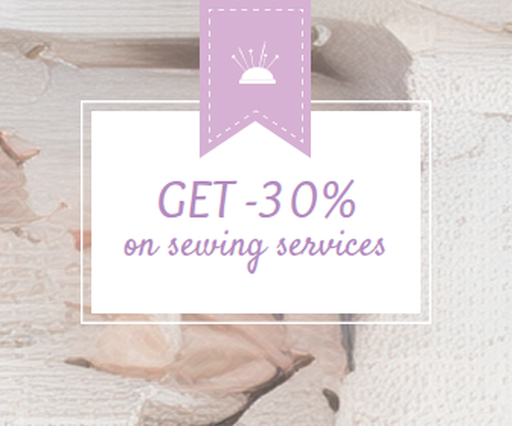} &
      \frfig{.32}{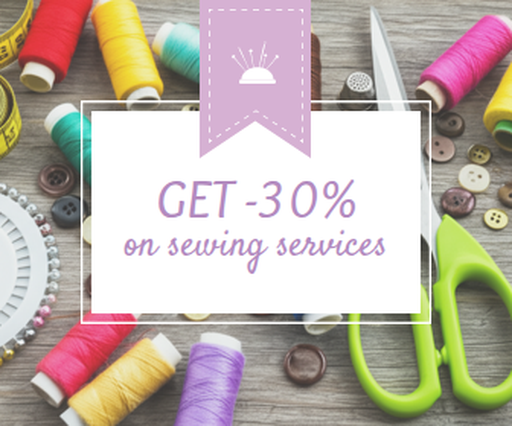} &
      \frfig{.32}{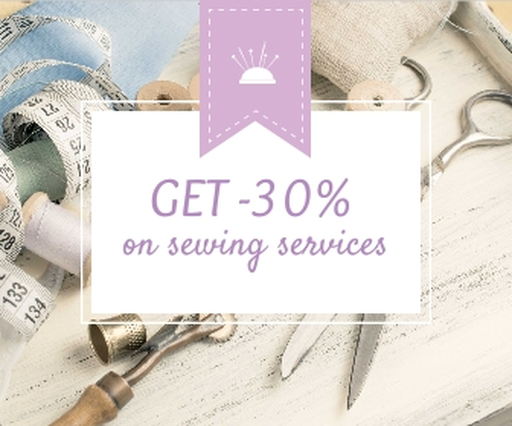} \\
      \multicolumn{3}{c}{\textit{Add sewing-related items (...) around the text in the background.}}
    \end{tabular} }
    \caption{\looseness=-1
    Qualitative results for instruction-guided completion with caption generation. The second column shows the result of using the external text-to-image model~\cite{Gemini} to generate the image based on the predicted caption.}
    \label{fig:instructed_completion_caption}
    \Description{This figure compares methods for adding sewing-related items behind a discount banner. The second image uses a text-to-image model to generate content based on a predicted caption. The ground truth shows various sewing tools placed around the text.}
\end{figure}

Given the recent success of text-to-image (T2I) models in generating high-quality images from text prompts, we also introduce a variant of our model, \emph{Instruct-MarkupDM$^\ast$}, which generates additional captions for image elements to leverage external T2I models. \Cref{fig:instructed_completion_caption} shows the qualitative results with and without using T2I. Without T2I, the model produces vague and unclear objects, whereas with T2I, the images are more detailed and better aligned with the instructions. This result demonstrates that our model benefits from recent T2I models to generate high-quality images. While these models may struggle with images requiring transparency or extreme aspect ratios, our image tokenizer can handle these needs. Our findings suggest that external T2I models can compensate for our model's limited image-generation capabilities while achieving instruction-guided completion within editable, structured graphic design templates.

\setlength{\tabcolsep}{1.5pt}
\setlength{\tabcolsep}{6pt}

\section{Limitations and Discussion}
We presented MarkupDM, a multimodal markup document model that integrates a large language model trained using the fill-in-the-middle objective and a specialized image tokenizer for images of variable sizes with transparency. By treating graphic designs as interleaved multimodal documents, our approach unifies text and image token generation within a single framework. Experimental results indicate that MarkupDM effectively completes various graphic design tasks, including attribute value prediction, image generation, and text insertion, while preserving the contextual relationships among design elements. Further extension to instruction-guided design completion demonstrates the flexibility of our approach, where it achieves competitive performance compared with state-of-the-art image editing models.

Despite these promising results, our approach has several limitations. First, the model still struggles to generate complex or highly detailed images. As shown in \cref{fig:instructed_completion_caption}, an external text-to-image model can generate primary image elements using predicted captions, but it is unclear whether it can produce decorative or background elements that visually harmonize with the surrounding content. Incorporating a more recent, powerful multimodal model such as Janus-Pro~\cite{ChenJanusPro2025} could solve this issue, although it lacks native fill-in-the-middle capabilities. Additionally, given the rapid progress in foundation models, agentic approaches to design automation are another promising direction~\cite{WangBannerAgency2025}.

\looseness=-1
Second, our model faces challenges in visually intricate compositional tasks that require nuanced spatial reasoning, such as layering multiple objects or maintaining aesthetic coherence across various elements. Enhancing the model's spatial understanding may require domain-specific training or dedicated spatial modules.

Finally, our current model primarily focuses on the generation of new elements rather than refining or editing existing elements in detail or creating entire documents from scratch. Although it can insert a missing component, full-fledged editing of already-placed objects (including detailed manipulations of shape and texture) remains outside its scope. Addressing these limitations in future work will likely involve larger and more diverse datasets. We hope our findings encourage further research on multimodal LLMs for design tasks and motivate the development of more sophisticated, user-driven design automation techniques.

\bibliographystyle{ACM-Reference-Format}
\bibliography{references}

\appendix

\section{Details on Our Crello-Instruct Dataset}
We extend the Crello dataset~\cite{YamaguchiCanvasVAE2021} (version 5.0.0) to create our dataset for instruction-guided graphic design completion, as described in \cref{sec:dataset}. We remove one element at a time from the original design templates to create partial designs. We then convert both the original and partial designs into images using the specialized renderer\footnote{\url{https://github.com/CyberAgentAILab/cr-renderer}} with a maximum size of 768 pixels and a light gray canvas color to prevent white elements from becoming invisible. Next, we employ the Qwen2.5-VL-7B-Instruct model~\cite{qwen25vl} to generate a brief editing instruction based on these images. Instructions are generated using the greedy decoding strategy with the following prompt:
\begin{tcolorbox}[
    colframe=gray!90,
    title={Prompt for instruction generation},
    fonttitle=\small,
    fontupper=\small,
    boxrule=1pt, %
    left=2.5pt, %
    right=2.5pt, %
    top=2.5pt, %
    bottom=2.5pt %
]
Picture 1 is the original image and Picture 2 is its edited version. Provide a very brief instruction for a designer to create the edited version based on the original image, using the imperative form.
\end{tcolorbox}
\noindent We then edit the generated instructions to remove reference phrases such as ``in Picture 1'' and ``in the edited version'' and replace phrases related to the light gray background with no color information to mitigate the influence of the generation pipeline.

\begin{figure}[t]
    \centering
    \includegraphics[width=\linewidth]{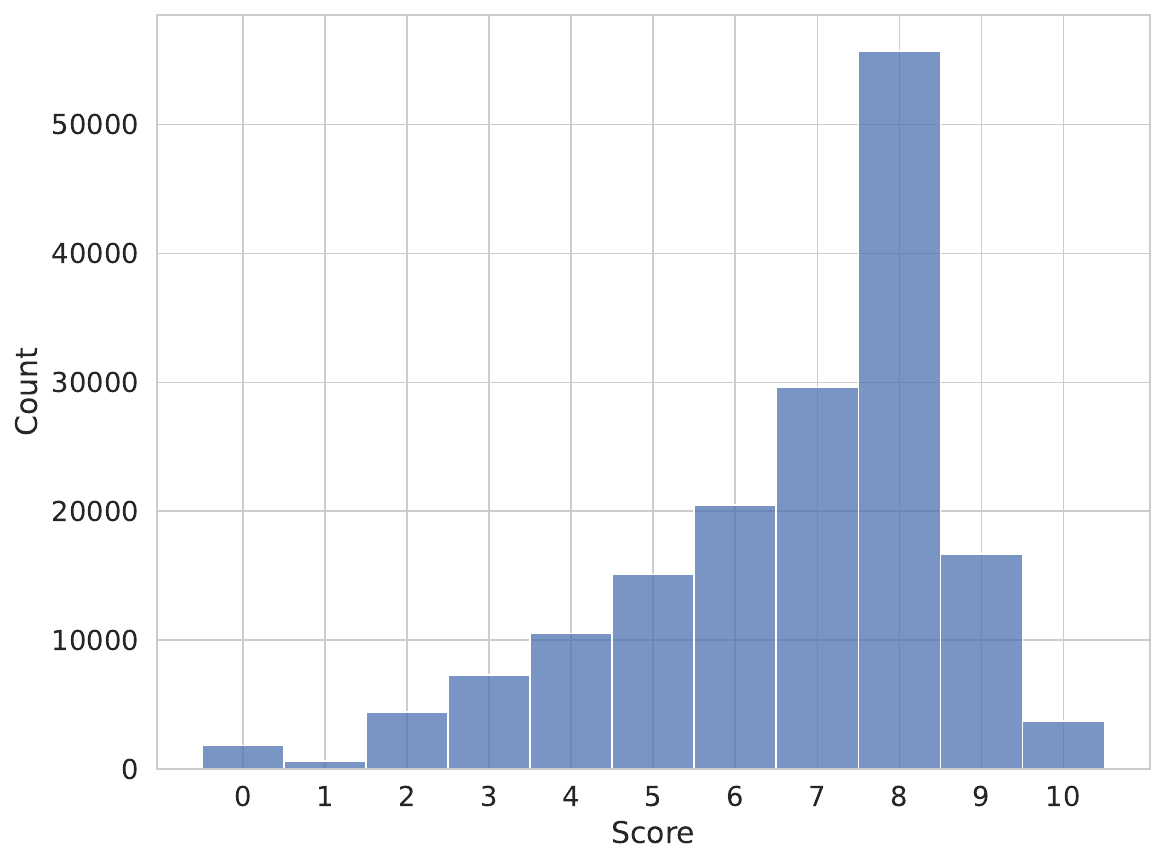}
    \caption{Instruction quality score distribution. The scores are generated by GPT-4o mini~\cite{GPT4oMini} and range from 0 (extremely inadequate) to 10 (excellent).}
    \label{fig:instruction_scores}
    \Description{TODO}
\end{figure}

Since the resulting instructions are often noisy, we filter out low-quality samples. First, we remove any samples where the instruction begins with ``Increase,'' ``Remove,'' ``Move,'' ``Decrease,'' ``Rotate,'' ``Reverse,'' or ``Update.'' Next, we use GPT-4o mini~\cite{GPT4oMini} to rate the quality of each triplet (instruction, partial design, and completed design) using the following prompt:
\begin{tcolorbox}[
    colframe=gray!90,
    title={Prompt for instruction rating},
    fonttitle=\small,
    fontupper=\small,
    boxrule=1pt, %
    left=2.5pt, %
    right=2.5pt, %
    top=2.5pt, %
    bottom=2.5pt %
]
The following editing instruction is machine-generated and may contain errors. Critically review the instruction along with the two images provided (first: pre-edit; second: post-edit). Identify any mistakes or inconsistencies, then assign a score from 0 (extremely inadequate) to 10 (excellent).
\newline\newline
Instruction: [INSTRUCTION]
\end{tcolorbox}
\noindent We filter out samples with a score of 5 or lower. The score distribution is shown in \cref{fig:instruction_scores}. As a result, we remove 32.5\% of the samples through first-word filtering, 15.9\% through rating-based filtering, and 0.9\% due to other factors such as GPT API errors. The final dataset contains 103,917 samples for training, 9,839 for validation, and 11,350 for testing.

Additionally, we generate a caption for each non-textual element in the dataset to help the model understand the image content. We use the same Qwen2.5-VL-7B-Instruct model~\cite{qwen25vl} with a simple prompt: ``Describe this image concisely.'' We observed that the model frequently outputs the phrase ``mathematical expression'' for images with little semantic content, such as decorative elements, and we remove those captions. We also strip redundant prefixes like ``The image is'' and ``The image shows'' from the captions.

\section{Implementation Details}
We build our RGBA tokenizer by finetuning a pre-trained RGB tokenizer, as described in \cref{sec:image_tokenizer,sec:image_tokenizer_setup}.
We train our tokenizer in 100,000 steps using mixed precision training using bfloat16 a batch size of 8 with a single NVIDIA L4 GPU.
We use the Adam optimizer with a learning rate of $1 \times 10^{-5}$. 
We resize the input images to $256 \times 256$ pixels.
The training takes approximately 2 days to complete.

We build our MarkupDM by extending pre-trained LLMs, as described in \cref{sec:MarkupDM,sec:markupdm_setup}.
We train our model in 100,000 steps with a single A100 80GB GPU with several techniques for efficient training, including mixed precision training using bfloat16, gradient checkpointing, and the Flash Attention 2.
We use the Adam optimizer with a learning rate of $5 \times 10^{-5}$ and a constant schedule.
The fill-in-the-middle (FIM) transformation is applied with a probability of 0.9 during training.
Specifically, we use the context-level FIM with the token-level span selection in the prefix-suffix-middle format~\cite{BavarianEfficient2022}.
The training the model using StarCoderBase-7B takes approximately 3 days to complete.

We train our instruction-tuned model, Instruct-MarkupDM, in 100,000 steps with the same hyperparameters as the base model. The document representation for this model starts with the instruction, followed by the prefix, suffix and middle parts of the original document. We compute the loss only for the middle part, which is the target to be filled in. We show an example document representation in the following:
\begin{tcolorbox}[
    colframe=gray!90,
    title=Multimodal markup document with an instruction,
    fonttitle=\small,
    fontupper=\small,
    boxrule=1pt, %
    left=2.5pt, %
    right=2.5pt, %
    top=2.5pt, %
    bottom=2.5pt %
]
\footnotesize
\begin{Verbatim}[commandchars=\\\{\}]
\textcolor{spec}{[bos]}
Change the background color of the image to dark blue.
\textcolor{fim}{[fim_prefix]}
<svg xmlns="http://www.w3.org/2000/svg" viewBox="0 0 419 298 "widt
h="419" height="298">
\textcolor{fim}{[fim_suffix]}
<text font-family="Montserrat" font-size="30" font-weight="bold" f
ill="rgba(255, 255, 255, 1)" x="32" y="81">FAMILY</text>...</svg>
\textcolor{fim}{[fim_middle]}
<image href="\textcolor{spec}{[boi]}360\textcolor{spec}{[sep]}260\textcolor{spec}{[sep]}\textcolor{img}{[img:1][img:42][img:3]}...\textcolor{spec}{[eoi]}"
x="-9" y="-9" width="436" height="315"/>
\textcolor{spec}{[eos]}
\end{Verbatim}
\end{tcolorbox}

\section{Details on Our Experiments and Additional Results}
We convert structured graphic design documents to SVG-based format, as described in \cref{sec:our_representation,sec:markupdm_setup}.
We show in \cref{tab:tags_and_attributes} the element tags and their corresponding attributes used in the experiments.
For some experiments, we also use the custom attribute of \emph{caption} for the image elements, which is used to store the description of the image.

\begin{table}[h]
    \centering
    \caption{Element tags and their corresponding attributes used in our experiments. Please refer to the SVG specification for details.}
    \label{tab:tags_and_attributes}
    \begin{tabular}{lL{6cm}}
    \toprule
    Element tag & Attribute \\
    \midrule
    svg & xmlns, viewBox, width, height \\[2mm]
    image & href, x, y, width, height, transform, opacity \\[2mm]
    text & x, y, fill, font-family, font-size, font-weight, font-style, text-anchor, letter-spacing, transform, opacity \\
    \bottomrule
    \end{tabular}
\end{table}

For graphic design completion, we use the top-$p$ sampling with $p\!=\!0.9$ for generation. We set the maximum number of new tokens to 10 for attribute value completion, 50 for text completion, 280 for image completion (256 for the image tokens, 4 for the special tokens, and 20 for the image width and height), and 350 for instruction-guided element completion.

\begin{figure*}[t]
    \centering
    \includegraphics[width=.92\linewidth]{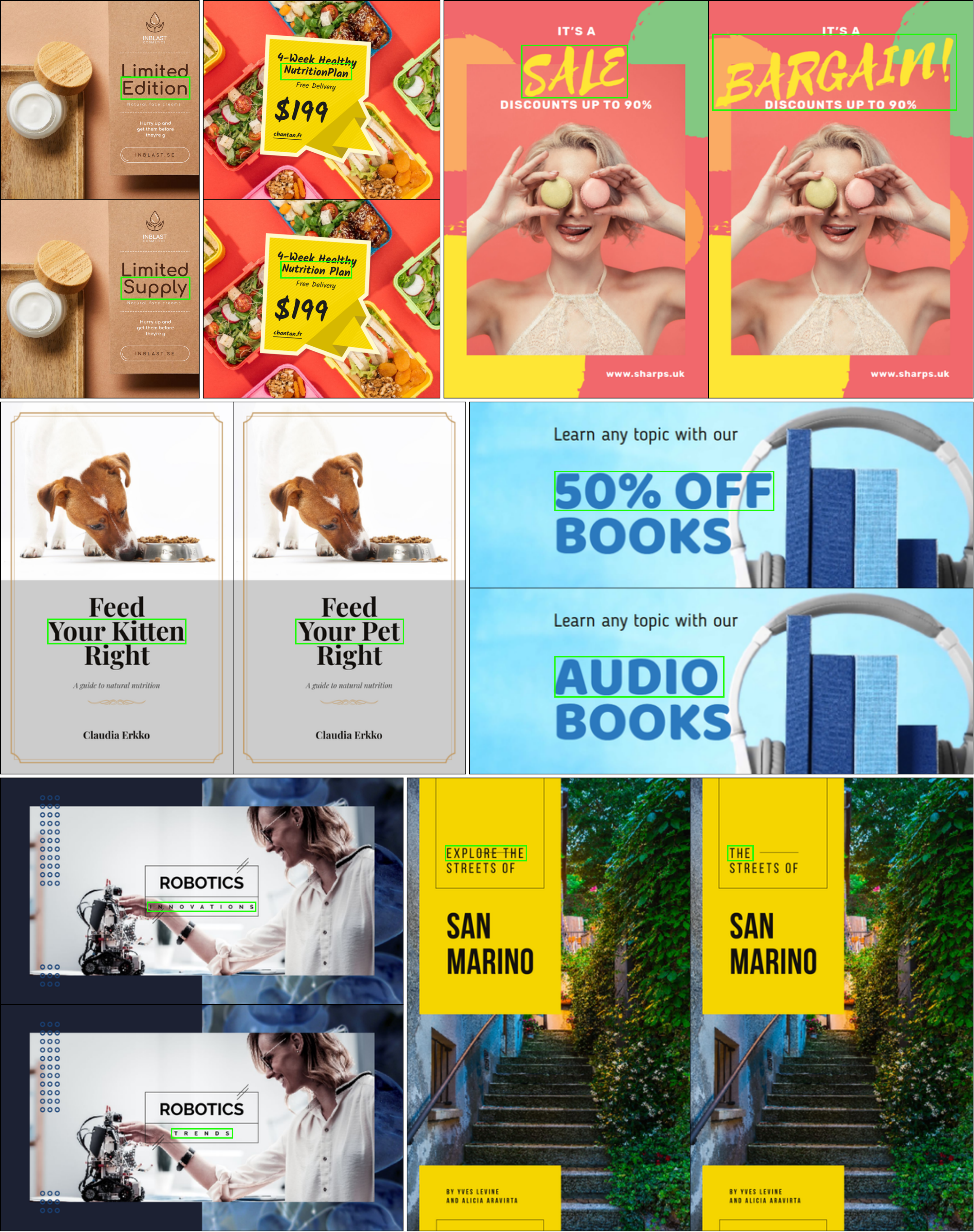}
    \caption{Text completion results. Each pair shows the predicted completion and the original design from left to right or top to bottom. The green boxes indicate the target text.}
    \label{app:fig:text_completion}
    \Description{TODO}
\end{figure*}

\begin{figure*}[t]
  \centering
  \includegraphics[width=.94\linewidth]{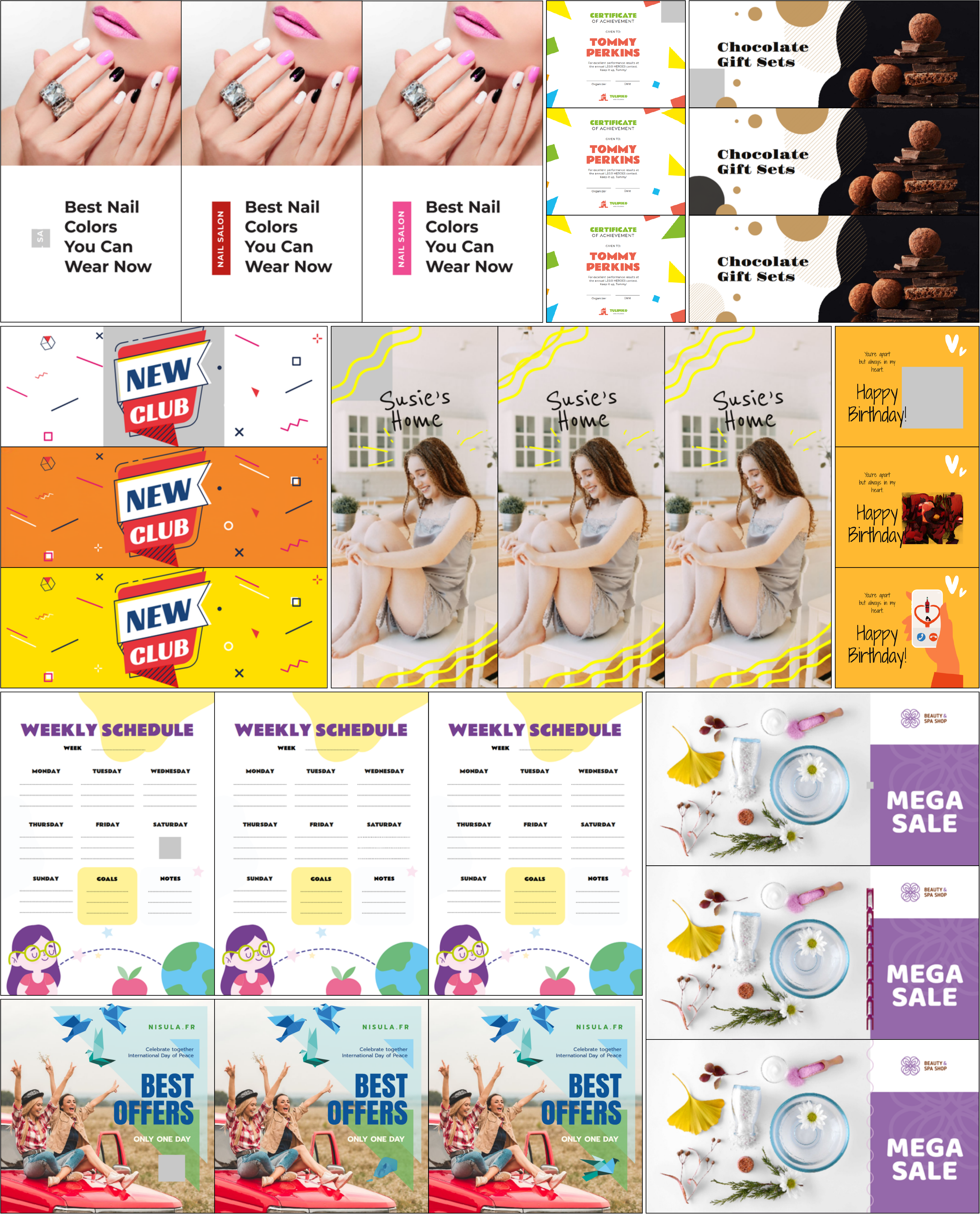}
  \caption{Image completion results. Each triplet shows the input, the predicted completion, and the original design from left to right or top to bottom. The gray squares indicate the target image elements to be completed.}
  \label{app:fig:image_completion}
  \Description{TODO}
\end{figure*}

We provide additional image reconstruction results in \cref{app:fig:image_reconstruction}, text completion results in \cref{app:fig:text_completion}, and image completion results in \cref{app:fig:image_completion} by MarkupDM using StarCoderBase-7B. We also provide additional qualitative results for instruction-guided graphic design completion in \cref{app:fig:instructed_completion}.

\setlength{\tabcolsep}{0pt}

\newcommand{\figqa}[1]{
    \frame{\includegraphics[width=.192\linewidth]{figures/supplementary/image_reconstruction/#1.png}}
}
\newcommand{\figrowa}[1]{
    \figqa{ldm-vq-f16_pred_#1}&
    \figqa{ldm-vq-f16-ft-part_pred_#1}~~&
    \figqa{ldm-vq-f16-ft-part_pred_rembg_#1}&
    \figqa{ldm-vq-f16-ft-part-rgba_pred_#1}&
    \figqa{target_#1}\\[-1.5mm]
}

\begin{figure*}[t]
    \centering
    \begin{tabular}{ccccc}
        \multicolumn{2}{c}{RGB images} & \multicolumn{3}{c}{RGBA images} \\
        LDM-VQ~\cite{RombachHighresolution2022} & Ours-RGB & Ours-RGB + Rembg~\cite{GatisRembg2020} & Ours & Original \\
        \figrowa{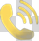}
        \figrowa{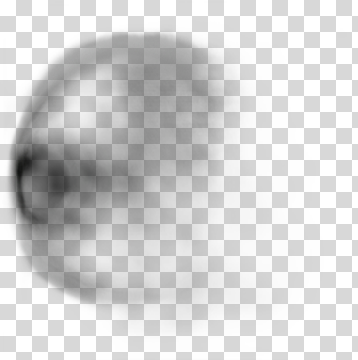}
        \figrowa{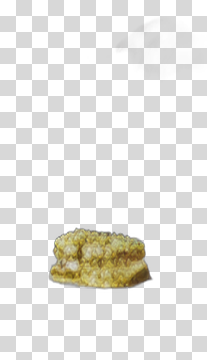}
        \figrowa{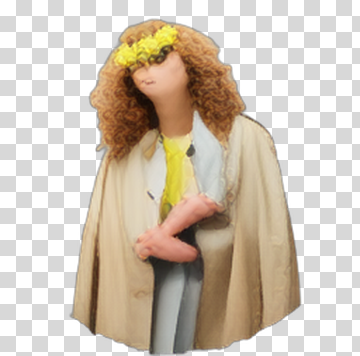}
        \figrowa{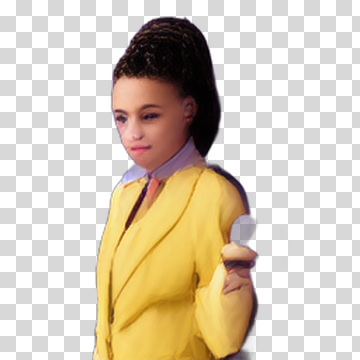}
    \end{tabular}
    \caption{Additional image reconstruction results. Our tokenizer ($f\!=\!16$) can encode RGBA images so that they can be reconstructed with high fidelity.
    Reconstructing human faces is still challenging for our tokenizer, but it may be alleviated by using finer scaling factors or additional losses for faces.}
    \label{app:fig:image_reconstruction}
    \Description{TODO}
\end{figure*}

\setlength{\tabcolsep}{6pt}
\renewcommand{\arraystretch}{1}

\setlength{\tabcolsep}{0pt}
\setlength{\tabcolsep}{6pt}

\setlength{\tabcolsep}{1.5pt}
\newcommand{\figrow}[2]{
\frfig{#1}{supplementary/instructed_completion/#2_input.png} &
\frfig{#1}{supplementary/instructed_completion/#2_HQ-Edit_pred.png} &
\frfig{#1}{supplementary/instructed_completion/#2_Tuned-HQ-Edit_pred.png} &
\frfig{#1}{supplementary/instructed_completion/#2_Gemini_pred.png} &
\frfig{#1}{supplementary/instructed_completion/#2_MarkupDM_pred.png} &
\frfig{#1}{supplementary/instructed_completion/#2_HQ-Edit_gt.png}
}
\begin{figure*}[t]
    \centering
    \begin{tabular}{cccccc}
      Input & HQ-Edit~\cite{HuiHQEdit2025} & HQ-Edit~\cite{HuiHQEdit2025} + FT & Gemini 2.0 FE~\cite{Gemini} & Instruct-MarkupDM & Ground Truth \\

      \figrow{.16}{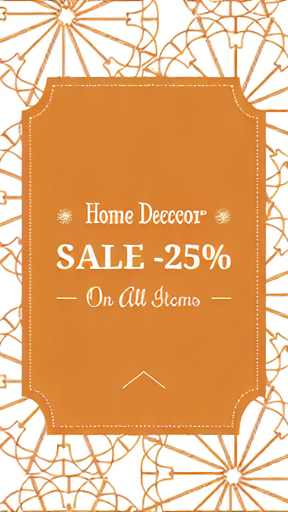} \\
      \multicolumn{6}{c}{\textit{Add ``Home Decor'' above ``SALE -25\%''.}} \\[6pt]

      \figrow{.16}{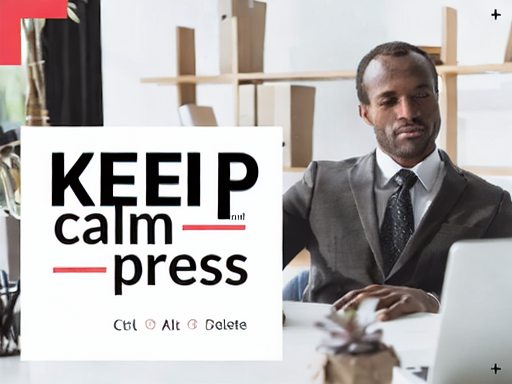} \\
      \multicolumn{6}{c}{\textit{Replace ``Calm Press'' with ``Keep Calm Press'' in the text box.}} \\[6pt]

      \figrow{.16}{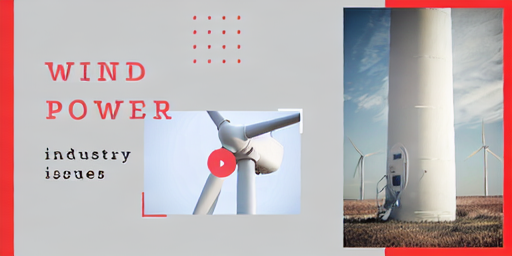} \\
      \multicolumn{6}{c}{\textit{Add a red vertical rectangle to the right side of the image, extending from the top to the bottom.}} \\[6pt]

      \figrow{.16}{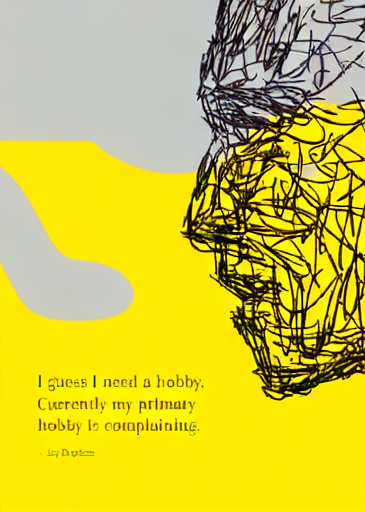} \\
      \multicolumn{6}{c}{\textit{Change the background color of the image to yellow.}} \\[6pt]

      \figrow{.16}{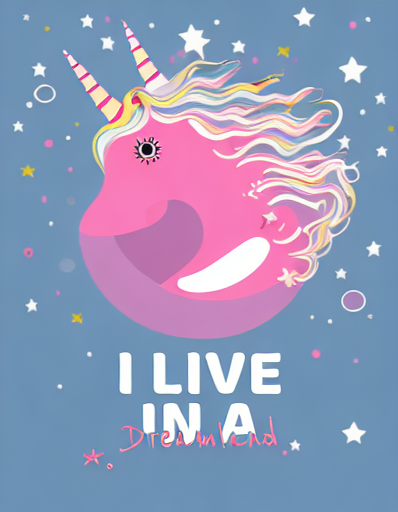} \\
      \multicolumn{6}{c}{\textit{Add a unicorn illustration with stars around it above the text.} (failure case)}

    \end{tabular}
    \caption{Qualitative comparison for instruction-guided graphic design completion.}
    \label{app:fig:instructed_completion}
    \Description{TODO}
\end{figure*}
\setlength{\tabcolsep}{6pt}

\end{document}